%% file: main.tex
\documentclass[runningheads]{llncs}
\usepackage[T1]{fontenc}
\usepackage{graphicx}
\usepackage[hidelinks]{hyperref}
\usepackage{CJKutf8}
\usepackage{amsmath,amssymb}
\usepackage{url}
\usepackage{booktabs}
\usepackage{siunitx}
\usepackage[font=small,labelfont=bf]{caption}
\usepackage{threeparttable}
\usepackage{multirow}
\usepackage[dvipdfmx]{xcolor}

\begin{document}
\begin{CJK}{UTF8}{min}
\title{Effects of Transcript Compression on LLM-based Medical Misinformation Detection in Japanese YouTube Videos}
\titlerunning{Transcript Compression for Medical Misinformation Detection}
%
%
\author{Yuya Wake \and
Sho Tsugawa \and
Toshiyuki Amagasa}
\authorrunning{Y. Wake et al.}
%
\institute{
University of Tsukuba,
1-1-1 Tennodai, Tsukuba, Ibaraki 305-8573, Japan\\
\email{s2630199@u.tsukuba.ac.jp}
}
\maketitle              
\begin{abstract}
Large language models (LLMs) are increasingly used to assess long-form medical videos,
but their effectiveness may depend on whether transcripts
are provided in full or compressed through summarization, retrieval, or claim screening.
This study examines how such
transcript compression affects LLM-based veracity classification of Japanese
medical YouTube videos. We compare four transcript input designs: full
transcripts, LLM-generated summaries, RAPTOR-based retrieval-augmented
generation (RAG), and Screening, which extracts candidate medical and
health-related sentences. Using 74 long-form videos labeled as Real or Fake,
we evaluate classification performance and analyze linguistic changes using
J-LIWC, hedge expressions, and institutional or technical terms. The
full-transcript Baseline achieved the best performance, whereas all compressed
inputs increased false negatives, meaning that Fake videos were more likely
to be misclassified as Real. Summary caused the largest performance drop,
while Screening performed best among the compressed inputs but still omitted
many medically relevant sentences. Linguistic analyses showed that these
errors were not explained by a simple increase in certainty. Instead, Summary
reduced affective, social, temporal, cognitive, and conversational cues, while
Summary and RAG made institutional and technical terms more salient. These
findings suggest that transcript compression can represent Fake videos as
more coherent and authoritative inputs, thereby weakening cues needed for
misinformation detection.
\footnote{All code, prompts, queries, and analysis resources, including the manually reviewed dictionaries, are publicly available at
\url{https://github.com/yuya-wake/ASONAM2026}.}
\end{abstract}

\section{Introduction}
\input{section/1_intro}

\section{Related Work}
\input{section/2_related}

\section{Dataset of Japanese-language Medical YouTube Videos}
\input{section/3_dataset}

\section{Comparison of Input Designs for LLM-based Veracity Classification}
\input{section/4_preliminary}

\section{Analysis of Compression-Induced Linguistic Changes}
\input{section/5_experiment}

\section{Discussion}
\input{section/6_discussion}

\section{Conclusion and Future Work}
\input{section/7_conclusion}

\section*{Acknowledgments}
\input{section/8_acknowledgements}

%
%
\bibliographystyle{splncs04}
\bibliography{bib/references}

\end{CJK}
\end{document}

%% file: section/1_intro.tex
Medical and health misinformation on video-sharing platforms
can cause tangible harm by encouraging poor treatment decisions,
fostering vaccine hesitancy, and delaying professional care
\cite{do2022infodemics,li2022youtube}.
YouTube is particularly important in this context
because it is widely used as a source of health information,
and health-related videos on the platform may contain inaccurate,
misleading, or low-quality medical content
\cite{madathil2015healthcare,osman2022youtube}.
These risks highlight the need for methods
that can detect medical and health misinformation on YouTube.

LLM-based veracity classification has emerged as a promising approach
for identifying medical misinformation in videos~\cite{khalil2025evaluating},
but its effectiveness depends critically on how long textual inputs
are represented and selected for the model~\cite{li2024long,liu2024lost}.
Full-transcript input preserves the maximum amount of textual information,
but transcripts are often lengthy, repetitive, and computationally costly.
In contrast, summarization, retrieval-augmented generation (RAG)~\cite{li2024long},
and claim screening can reduce input length
by rewriting, retrieving, or selecting parts of the transcript
\cite{lewis2020retrieval,jiang2023llmlingua,jiang2024longllmlingua}.
However, because these methods do not preserve the full transcript,
they may remove or alter contextual and linguistic cues
that are important for veracity classification.

Although prior studies have examined how LLMs process long contexts
\cite{liu2024lost,li2024long}
and how input compression can reduce inference cost or improve long-context processing
\cite{jiang2023llmlingua,jiang2024longllmlingua},
much of this work has focused on tasks such as question answering,
where the main objective is to preserve or retrieve answer-bearing information
from long inputs.
Medical-video veracity classification differs from such tasks because it requires
judging how multiple claims, explanations, testimonials, hedge expressions,
emotional appeals, and institutional or technical terms jointly shape video
credibility.
Even small omissions or changes in these cues may affect the classification outcome.
Thus, it remains unclear how transcript compression affects LLM-based veracity classification
and what linguistic changes may underlie misclassifications.

This study investigates how transcript input design affects
LLM-based veracity classification of Japanese-language medical YouTube videos.
We manually labeled the 74 long-form medical YouTube videos included
in the final analysis as Real or Fake and evaluated
whether an LLM could correctly classify their veracity
under different transcript input designs.

Our main contributions are threefold.
First, we compare four transcript input designs under the same classification LLM and prompt,
thereby isolating the effect of transcript representation.
Second, we show that transcript compression increases false negatives
compared with full-transcript input.
Third, we analyze compression-induced linguistic changes using J-LIWC~\cite{igarashi2022development},
hedge expressions, and institutional or technical terms.

%% file: section/2_related.tex
Medical and health-related misinformation detection
has traditionally been studied
using multimodal and contextual signals,
including textual content, metadata, user reactions,
diffusion structures, and visual information
\cite{hou2019towards,zhou2020recovery,shang2025multitec}.
For video-sharing platforms,
misinformation detection is more complex
because relevant information is distributed across titles,
descriptions, transcripts, visual content, and engagement metrics
\cite{madathil2015healthcare,osman2022youtube}.
This complexity is especially pronounced in medical videos,
which often combine treatment claims, diagnostic advice,
preventive recommendations, and personal testimonials
within the same video
\cite{kang2026quality}.

Recent studies have explored LLMs for fact-checking and misinformation
detection because they can interpret claims, generate explanatory judgments, and handle complex textual evidence
\cite{vykopal2024generative}.
However, many misinformation detection studies have focused on shorter textual units,
such as individual claims
\cite{sarrouti2021evidence},
news articles~\cite{zhou2020recovery},
or social media posts~\cite{hossain2020covidlies}.
Although prior work has examined misinformation
detection from YouTube transcripts~\cite{christodoulou2023identifying}
and LLM-based assessment of medical YouTube content~\cite{khalil2025evaluating},
long-form health videos pose additional challenges.
They often combine factual claims with narrative and persuasive cues,
such as authority signals, personal stories, emotional appeals,
and critiques of mainstream medicine ~\cite{caulfield2019health}.
Because such cues may be distributed across long transcripts, and because LLMs
can be sensitive to the location and selection of relevant information in long
contexts~\cite{liu2024lost,qi2024long2rag},
it remains unclear how effectively LLMs can classify the veracity of long-form medical video transcripts.

In addition to misinformation detection, this study is also related to
long-context processing and input compression for LLMs.
To process long documents with LLMs,
prior work has examined full-document input, summarization,
and retrieval of relevant passages with RAG.
Claim extraction can also be viewed as an extractive form of compression
that selects task-relevant statements before classification.
Prior studies report that full-context approaches can outperform RAG in question-answering tasks,
while longer inputs can also degrade model performance
\cite{li2024long}.
Thus, both information preservation and input reduction must be considered
when designing transcript inputs for LLM-based classification.

Despite these advances, the effects of transcript input design
on LLM-based veracity classification remain unclear.
Medical-video transcripts contain heterogeneous cues, including claims,
explanatory flow, testimonials, hedges, emotional cues, and
institutional and technical terms.
Prior studies suggest that
stylistic and lexical cues,
such as emotional expressions, sensational language, source-related features,
and vocabulary associated with expertise or institutions, can influence misinformation detection
\cite{liu2024emotion,wu2024fake}.
These cues may be
unevenly preserved, removed, or foregrounded by compression.
Summarization may rewrite spoken discourse,
RAG may retrieve only selected fragments,
and claim extraction may omit narrative or persuasive context.
Therefore, it remains unclear which input design
is most effective and what linguistic changes may underlie misclassifications.

%% file: section/3_dataset.tex
This section describes the construction process of a Japanese-language
medical YouTube video dataset designed to evaluate how transcript
compression affects LLM-based veracity classification.
Specifically, we explain the target topics, video collection process, label definitions, transcript generation procedure, and selection criteria for the videos included in the analysis.
We focused on five topics in which medical or health-related misinformation is frequently observed:
alternative medicine, COVID-19, cancer treatment, HPV vaccines, and monosodium glutamate.
Videos were collected through crowdsourcing.
Ten crowd workers collected videos for these topics based on the definitions of Fake and Real videos described below, resulting in 921 videos in total.
Metadata, including titles and descriptions, were obtained via the YouTube Data API through the YouTube Researcher Program.
\footnote{\url{https://research.youtube/} (Accessed: 2025-11-26).}
To comply with platform policies, protect privacy, and avoid the risk of defamation, we do not release the raw video data, video metadata, or assigned labels.

Video labels were assigned based on both medical validity and the potential health risks
that could arise if viewers believed or acted on the video content.
Opinion videos were those whose main content consisted of subjective personal opinions
and for which absolute veracity judgments were difficult.
Real videos were medically valid videos associated with a low risk of health harm
if their content was believed, whereas Fake videos contained false or misleading information
that could lead to health harm if believed. Opinion videos were excluded from the analysis,
and binary classification was conducted using only Real and Fake videos.
All labels were assigned by one author.

To represent video content as LLM input,
we generated transcripts from audio
using Whisper
\footnote{We used the large-v2 model.}
\cite{radford2023robust}
after downloading the videos with permission from the
YouTube Researcher Program.
Because the transcripts contained Japanese character-conversion errors,
including incorrect kanji substitutions, we used Gemini
\footnote{We used gemini-1.5-flash.}
\cite{team2023gemini}
to correct such transcription errors.
We also removed URL strings from video descriptions
to avoid unnecessary increases in input length.

To ensure that transcript compression was meaningfully required,
we retained only videos whose transcripts contained at least 10,000 Japanese characters.
For relatively short transcripts, the computational burden of full-transcript input is limited,
and the potential information loss caused by compression may outweigh its benefits.
Therefore, we restricted the analysis to long-form videos for which transcript compression
is more likely to be practically necessary.
The final analysis set consisted of 74 videos: 10 Real videos and 64 Fake videos.


%% file: section/4_preliminary.tex
This section compares four transcript input designs to examine
how transcript compression affects LLM-based veracity classification performance.

The four designs are the Baseline condition, Summary condition, RAG condition, and Screening condition.

\subsection{Input Design for LLMs}

To isolate the effect of transcript input design,
we evaluated the following four transcript input designs using the same classification LLM,
classification prompt, video title, and video description.

\begin{itemize}
    \item {\textit{Baseline}: The Baseline condition provides the full transcript without compression and serves as the full-transcript information reference condition. All samples fit within the context length of the classification LLM.}
    \item {\textit{Summary}: The Summary condition uses an LLM-generated summary of the transcript and represents generative input compression. We used Qwen2~\cite{team2024qwen2} as the summarization model based on a preliminary model selection experiment.\footnote{We used \texttt{qwen2-7b-instruct-q5\_k\_m.gguf}.}}
    \item {\textit{RAG}: The RAG condition retrieves relevant transcript contexts using RAPTOR~\cite{sarthi2024raptor} and represents retrieval-based compression. RAPTOR recursively clusters and enables retrieval from both local and global levels of the transcript.}
    \item {\textit{Screening}: The Screening condition filters the transcript to medical and health-related sentences and their surrounding local contexts, representing task-oriented extractive compression. An LLM extracts medical or health-related sentences from sentence-based chunks while preserving the original wording. No fallback case occurred.}
\end{itemize}

\subsection{Experimental Settings and Evaluation Metrics}

We used the same Japanese-compatible local LLM for
all classification experiments to ensure that performance differences
were attributable to transcript input design rather than model choice.
Specifically, we used the non-quantized FP16 version of
\nolinkurl{cyberagent/calm3-22b-chat}
available on Hugging Face as the classification model.
The same classification prompt was also used across all conditions.
\footnote{The prompt we used is available in \url{https://github.com/yuya-wake/ASONAM2026}}.
The classification LLM was queried with the video title, video description, and transcript input for each condition.
The LLM was prompted to classify the video as Real or Fake based on the provided information.
We fixed the temperature at 0 for all classification queries to minimize randomness in the LLM outputs.

We evaluated each condition using Accuracy, weighted F1-Score, Matthews Correlation Coefficient (MCC)
\cite{matthews1975comparison},
and the confusion matrix.
Because the
dataset is imbalanced, we report MCC in addition to Accuracy.
Fake videos were treated as the positive class; therefore,
a false negative refers to a Fake video misclassified as Real.

\subsection{Classification Results}

Table~\ref{tab:overall_metrics} compares the classification
performance for each input condition. In this table, Fake videos are treated
as the positive class and Real videos as the negative class. Therefore, TP
indicates Fake videos correctly classified as Fake, FN indicates Fake videos
misclassified as Real, TN indicates Real videos correctly classified as Real,
and FP indicates Real videos misclassified as Fake.

\begin{table*}[tb]
    \centering
    \caption{Classification performance of each input design}
    \label{tab:overall_metrics}
    \begin{tabular}{lrrrrrrr}
        \toprule
        Condition & Accuracy & F1-Score & MCC & TN & FN & FP & TP \\
        \midrule
        Baseline  & \textbf{0.905} & \textbf{0.891} & \textbf{0.524} & 4 & \textbf{1}  & 6 & \textbf{63} \\
        Summary   & 0.541 & 0.608 & 0.206 & \textbf{8} & 32 & \textbf{2} & 32 \\
        RAG       & 0.770 & 0.800 & 0.365 & 7 & 14 & 3 & 50 \\
        Screening & 0.824 & 0.838 & 0.391 & 6 & 9 & 4 & 55 \\
        \bottomrule
    \end{tabular}
\end{table*}

Table~\ref{tab:overall_metrics} shows that the full-transcript
Baseline achieved the best overall performance, whereas all compressed input
designs increased false negatives. The Baseline condition achieved an Accuracy
of 0.905, a weighted F1-Score of 0.891, and an MCC of 0.524. It also
produced only one false negative, indicating that almost all Fake videos were
correctly classified as Fake.

Overall, the full-transcript Baseline achieved the best performance
and produced the fewest false negatives,
although it still misclassified several Real videos as Fake. By contrast,
all compressed inputs reduced overall performance and increased false negatives,
suggesting that compression may have removed cues necessary for detecting Fake videos.
This motivates the following analysis of compression-induced linguistic changes.

%% file: section/5_experiment.tex
This section analyzes compression-induced linguistic changes
as possible factors underlying the increase in false negatives observed in Section 4.
For each analytical perspective, we describe the measurement procedure and report the corresponding results.
We focus on four perspectives: tentativeness and certainty, formal and conversational hedges,
institutional and technical terms, and the extraction performance of Screening.

\subsection{Common Analysis Procedure}

To compare linguistic features across inputs of different lengths,
we normalized each feature by the total number of words in the corresponding input.
For each video and each linguistic feature,
\(x_{\mathrm{baseline}}\) denotes the occurrence rate of the feature
in the Baseline input, whereas \(x_{\mathrm{compressed}}\) denotes
the occurrence rate of the same feature in a compressed input.
We then compared the Baseline input with each compressed input using the following difference:
\[
\Delta = x_{\mathrm{baseline}} - x_{\mathrm{compressed}}.
\]
A positive \(\Delta\) indicates that the feature decreased after compression,
whereas a negative \(\Delta\) indicates that it increased after compression.
We tested whether the median of \(\Delta\) differed from zero
using the Wilcoxon signed-rank test.
Because multiple linguistic features were tested,
we controlled the False Discovery Rate (FDR), that is,
the expected proportion of false positives among statistically significant results,
using the Benjamini-Hochberg procedure.

\subsection{Changes in Tentativeness and Certainty}

We first examined tentativeness and certainty
to test whether compression made the input appear less hesitant or more assertive.
This analysis was motivated by
our preliminary observation that compression, especially summarization, often
removed vague or hesitant expressions and reconstructed spoken discourse into
more assertive explanatory text. We analyzed the \textit{tentat} and
\textit{certain} categories in J-LIWC~\cite{igarashi2022development} after segmenting each text using
MeCab~\cite{kudo2004applying}, IPADIC, and the J-LIWC user dictionary.

Table~\ref{tab:certain_tentat_wilcoxon} shows the differences
in the occurrence rates of words classified into the J-LIWC
\textit{certain} and \textit{tentat} categories between the Baseline
and each compressed input condition.
Here, \(x_{\mathrm{baseline}}\) and \(x_{\mathrm{compressed}}\) denote
the occurrence rates of words in each J-LIWC category in the Baseline
input and the compressed input, respectively.
In the Summary condition, both \textit{certain} and \textit{tentat}
significantly decreased after compression.
By contrast, neither category showed a significant change in the RAG
or Screening conditions.
Thus, this pattern suggests that summarization did not simply make inputs more assertive;
rather, it broadly reduced both tentative and certain expressions.
Therefore, changes in tentativeness and certainty alone do not explain
why compressed inputs increased false negatives.

\begin{table}[t]
\centering
\begin{threeparttable}
\caption{Differences in \textit{certain} and \textit{tentat} across conditions}
\label{tab:certain_tentat_wilcoxon}
\begin{tabular}{llrrr}
\toprule
Condition & Feature & Compressed Rate & $\Delta$ & p-value \\
\midrule
\multirow{2}{*}{Summary}
  & certain & 0.0124 & 0.00372 & $3.52 \times 10^{-7}$ \\
  & tentat  & 0.0119 & 0.000646 & 0.00665 \\
\midrule
\multirow{2}{*}{RAG}
  & certain & 0.0117 & 0.000108 & 1.000 \\
  & tentat  & 0.00943 & 0.0000842 & 0.797 \\
\midrule
\multirow{2}{*}{Screening}
  & certain & 0.0104 & -0.000203 & 0.894 \\
  & tentat  & 0.00970 & 0.000181 & 0.575 \\
\bottomrule
\end{tabular}
\begin{tablenotes}[flushleft]
\footnotesize
\item Compressed Rate represents the occurrence rate of each feature
in the corresponding compressed input condition.
\(\Delta\) denotes the difference between the Baseline input and the compressed input
\((x_{\mathrm{baseline}} - x_{\mathrm{compressed}})\).
Positive \(\Delta\) values indicate lower rates after compression,
whereas negative values indicate higher rates after compression.
P-values are Wilcoxon signed-rank test results adjusted by the
Benjamini--Hochberg procedure to control the FDR.
\end{tablenotes}
\end{threeparttable}
\end{table}

For other J-LIWC categories,
Table~\ref{tab:other_significant_liwc} shows features that were significant
after FDR correction
and had relatively large absolute $\Delta$ values.
Summary significantly decreased multiple categories related to
affective, cognitive, temporal, and social processes.
RAG showed no significant J-LIWC changes after FDR correction,
whereas Screening showed a significant change only for verbs.

\begin{table}[tb]
\centering
\begin{threeparttable}
\caption{Other significant J-LIWC features with relatively large mean differences}
\label{tab:other_significant_liwc}
\begin{tabular}{llrrr}
\toprule
Condition & Feature & Compressed Rate & $\Delta$ & p-value \\
\midrule
\multirow{8}{*}{Summary}
  & \textit{affect} (affective processes)  & 0.0345 & 0.00942 & $9.85 \times 10^{-9}$ \\
  & \textit{posemo} (positive emotion)     & 0.0192 & 0.00691 & $8.13 \times 10^{-8}$ \\
  & \textit{cogproc} (cognitive processes) & 0.0755 & 0.00610 & $9.85 \times 10^{-9}$ \\
  & \textit{relativ} (relativity)          & 0.0624 & 0.00531 & $9.85 \times 10^{-9}$ \\
  & \textit{drives} (drives)               & 0.0551 & 0.00423 & $8.13 \times 10^{-8}$ \\
  & \textit{time} (time orientation)       & 0.0328 & 0.00341 & $8.75 \times 10^{-8}$ \\
  & \textit{social} (social processes)     & 0.0274 & 0.00301 & $9.83 \times 10^{-7}$ \\
  & \textit{negemo} (negative emotion)     & 0.0138 & 0.00203 & $1.60 \times 10^{-5}$ \\
\midrule
\multirow{1}{*}{Screening}
  & \textit{verb} (verbs)                  & 0.0534 & 0.00178 & 0.0305 \\
\bottomrule
\end{tabular}
\begin{tablenotes}[flushleft]
\footnotesize
\item Rate, \(\Delta\), and adjusted p-values are defined as in Table~\ref{tab:certain_tentat_wilcoxon}.
This table reports significant J-LIWC features other than \textit{certain} and \textit{tentat}.
\end{tablenotes}
\end{threeparttable}
\end{table}

\subsection{Changes in Formal and Conversational Hedges}

We next examined whether compression changed
not only the frequency of uncertainty expressions but also their style.
This analysis was motivated by the possibility that summaries may remove conversational hesitation
while retaining or introducing more formal uncertainty expressions.
We therefore analyzed conversational and formal
hedges. Hedges are expressions that weaken the certainty or assertiveness of an
utterance. Following Xiao~\cite{xiao2021hedge}, we constructed a list of candidate hedge expressions and
manually classified them into formal and conversational hedges (Table~\ref{tab:hedge_lexicon}).

\begin{table}[tb]
    \centering
    \caption{List of formal and conversational hedges, shown as Japanese expression (English gloss).}
    \label{tab:hedge_lexicon}
    \begin{tabular}{lp{0.72\linewidth}}
        \toprule
        Type & Hedge expressions \\
        \midrule
        Formal hedges &
        おそらく(\textit{probably}),
        可能性(\textit{possibility}),
        と考えられる(\textit{it is considered that}),
        とは限らない(\textit{not necessarily}),
        場合もある(\textit{in some cases}) \\
        Conversational hedges &
        たぶん(\textit{maybe/probably}),
        なんか(\textit{kind of}),
        気がする(\textit{I feel that}),
        かも(\textit{might}),
        かな(\textit{I wonder}),
        みたいだ(\textit{seems like}) \\
        \bottomrule
    \end{tabular}
\end{table}

Table~\ref{tab:hedge_formal_conversational} shows the differences in formal hedge rates,
conversational hedge rates,
and total hedge rates between the Baseline and each compressed input condition.
For each hedge category, $x_{\mathrm{baseline}}$ denotes
the occurrence rate of hedge expressions in the Baseline input,
whereas $x_{\mathrm{compressed}}$ denotes the occurrence rate
of hedge expressions in each compressed input.

Overall, transcript compression tended to reduce conversational hedges,
but this effect differed across input designs. In the Summary condition,
conversational hedge rates and total hedge rates significantly decreased
after compression, whereas formal hedge rates significantly increased.
This pattern suggests that summarization removed conversational uncertainty
while making formal uncertainty expressions relatively more salient.
Because the increase in formal hedges was smaller than the decrease in conversational hedges,
this change should be interpreted mainly as the removal of conversational uncertainty.
In the RAG condition, conversational and total hedge rates significantly decreased,
but formal hedge rates did not significantly change,
suggesting that RAG reduced conversational uncertainty without systematically
increasing formal hedging. In the Screening condition,
no hedge category showed a significant change after FDR correction,
suggesting that Screening altered hedge usage less than Summary or RAG.

\begin{table}[t]
\centering
\begin{threeparttable}
\caption{Changes in formal and conversational hedge rates across conditions}
\label{tab:hedge_formal_conversational}
\begin{tabular}{llrrr}
\toprule
Condition & Feature & Compressed Rate & $\Delta$ & p-value \\
\midrule
\multirow{3}{*}{Summary}
  & formal hedge rate         & 0.00523 & -0.00289 & $1.01 \times 10^{-7}$ \\
  & conversational hedge rate & 0.00896 & 0.0374   & $4.01 \times 10^{-13}$ \\
  & total hedge rate          & 0.0142  & 0.0346   & $3.13 \times 10^{-12}$ \\
\midrule
\multirow{3}{*}{RAG}
  & formal hedge rate         & 0.00282 & $-3.06 \times 10^{-6}$ & 0.592 \\
  & conversational hedge rate & 0.0311  & 0.00728                & $1.17 \times 10^{-9}$ \\
  & total hedge rate          & 0.0339  & 0.00728                & $1.17 \times 10^{-9}$ \\
\midrule
\multirow{3}{*}{Screening}
  & formal hedge rate         & 0.00295 & $-9.01 \times 10^{-5}$ & 0.688 \\
  & conversational hedge rate & 0.0318  & 0.000970               & 0.098 \\
  & total hedge rate          & 0.0347  & 0.000880               & 0.180 \\
\bottomrule
\end{tabular}
\begin{tablenotes}[flushleft]
\footnotesize
\item Rate, $\Delta$, and adjusted p-values follow the definitions in Table~\ref{tab:certain_tentat_wilcoxon}.
Formal hedge rate and conversational hedge rate represent the occurrence rates
of the manually defined formal and conversational hedge expressions, respectively.
Total hedge rate represents the combined occurrence rate of both hedge types.
\end{tablenotes}
\end{threeparttable}
\end{table}

\subsection{Foregrounding of Institutional and Technical Terms}

We also examined whether transcript compression foregrounded institutional
and technical terms in the compressed inputs.
This analysis was motivated by the possibility that compressed inputs,
especially summaries and retrieved contexts, may make public, professional,
or technical vocabulary more salient while reducing surrounding conversational context.
Such a shift may make medical content appear more authoritative, even when the
underlying claims are misleading.

Institutional and technical terms refer to words that connect medical
information to public systems, professional organizations, qualifications,
procedures, research knowledge, or socially legitimized frameworks of
judgment. Candidate terms were extracted using an LLM and then manually
reviewed by the author.
The prompts and resulting term list are included in the public repository for reproducibility.
For each input, we calculated both the overall
occurrence rate of institutional and technical terms and the occurrence rates
of individual terms.

Table~\ref{tab:institutional_terms} shows the institutional and technical term rates
that were significant after FDR correction.
The total rate significantly increased in both Summary and RAG,
indicating that institutional and technical terms were relatively
foregrounded in these compressed inputs. By contrast, Screening showed no significant change.
In Summary, however, the occurrence rates of ``論文'' (\textit{paper}) and ``病院'' (\textit{hospital}) significantly
decreased despite the increase in the total rate.
This indicates that the foregrounding of institutional
and technical terms reflected a shift in the overall vocabulary composition
rather than a uniform increase across all terms.

\begin{table}[t]
\centering
\begin{threeparttable}
\caption{Significant institutional and technical term rates across conditions}
\label{tab:institutional_terms}
\begin{tabular}{llrrr}
\toprule
Condition & Feature & Compressed Rate & $\Delta$ & p-value \\
\midrule
\multirow{3}{*}{Summary}
  & total rate & 0.00719 & -0.00239 & 0.0377 \\
  & 論文 (\textit{paper}) rate & 0.0000987 & 0.000199 & 0.000372 \\
  & 病院 (\textit{hospital}) rate & 0.000181 & 0.000167 & 0.0377 \\
\midrule
\multirow{1}{*}{RAG}
  & total rate & 0.00663 & -0.00203 & 0.0241 \\
\bottomrule
\end{tabular}
\begin{tablenotes}[flushleft]
\footnotesize
\item Rate, \(\Delta\), and adjusted p-values are defined as in Table~\ref{tab:certain_tentat_wilcoxon}.
Total rate represents the overall occurrence rate of institutional and technical terms.
\end{tablenotes}
\end{threeparttable}
\end{table}

\subsection{Supplementary Analysis: Extraction Performance of the Screening Condition}

The preceding analyses showed that Screening produced fewer compression-induced
linguistic changes than Summary and RAG, yet it still increased false negatives
compared with the Baseline. This raises the question of why classification
performance deteriorated even when the measured linguistic shifts were relatively
small. One possible explanation is that Screening may have omitted relevant
medical context during the sentence extraction stage. Therefore, we additionally
evaluated the extraction performance of the Screening condition.

Because Screening selects candidate medical and health-related sentences before
classification, this evaluation aimed to assess whether relevant medical context
was omitted at the extraction stage. We randomly selected eight videos, manually
annotated medical and health-related sentences, and calculated Precision, Recall,
and F1-score against the Screening outputs. Because one author conducted the
annotation, subjectivity remains a limitation.

The overall Precision, Recall, and F1-score were 0.777, 0.227, and 0.352,
respectively (Table~\ref{tab:screening_accuracy}).
The low-recall tendency was consistent across all eight sampled
videos, with per-video Recall ranging only from 0.119 to 0.472. Although
Precision was relatively high in most videos, one video showed substantially
lower Precision, indicating that extraction quality also varied depending on the
input. Overall, Screening tended to select relevant medical or health-related
sentences when it extracted them, but it missed many relevant sentences across
videos. This suggests that Screening may preserve some important claims while
omitting substantial surrounding medical context for veracity classification.

\begin{table}[tb]
\centering
\begin{threeparttable}
\caption{Overall extraction performance of the Screening Condition}
\label{tab:screening_accuracy}
\begin{tabular}{rrrrrr}
\toprule
TP & FP & FN & Precision & Recall & F1-Score \\
\midrule
192 & 55 & 653 & 0.777 & 0.227 & 0.352 \\
\bottomrule
\end{tabular}
\begin{tablenotes}[flushleft]
\footnotesize
\item TP, FP, and FN indicate true positives, false positives, and false negatives, respectively.
\end{tablenotes}
\end{threeparttable}
\end{table}


%% file: section/6_discussion.tex
This section discusses why full-transcript input outperformed compressed inputs,
how compression changed linguistic cues relevant to veracity classification,
and what limitations should be considered when interpreting the findings.

\subsection{Why Full-Transcript Input Outperformed Compression}

The central finding of this study is that full-transcript input
was more reliable and more effective than compressed input for LLM-based veracity classification
when the transcript fit within the context window and computational cost was acceptable.
The Baseline condition achieved the highest classification
performance, whereas all compressed input designs increased false negatives,
that is, cases in which Fake videos were misclassified as Real.

This finding is noteworthy because long inputs are not always assumed to be
beneficial.
Prior work has suggested that long contexts can make it difficult
for LLMs to attend to relevant information~\cite{liu2024lost},
and that input reduction can sometimes improve efficiency or performance
\cite{jiang2023llmlingua,jiang2024longllmlingua}.
In our setting, however, the
potential disadvantages of long input did not outweigh the disadvantages of
compression.
For medical video transcripts, relevant cues are distributed
across claims, testimonials, hedge expressions, emotional appeals, explanatory
flow, and technical and institutional terms. Full-transcript input preserved
these distributed cues more effectively than the compressed input designs.

The lower performance of compressed inputs does not imply that transcript
compression is inherently inappropriate.
Rather, it suggests that generic compression methods can reconstruct the input in ways
that weaken cues relevant to veracity classification.
Summary reduced affective, cognitive, temporal, social, and conversational cues,
while Summary and RAG foregrounded institutional and technical terms.
These changes may make the compressed representations of Fake videos appear more coherent
and authoritative than the original spoken discourse.
Therefore, future compression methods should not aim only to reduce
input length, but should preserve cues that are relevant to veracity
classification.

\subsection{Condition-specific Effects of Compression}

The compressed input designs degraded classification performance in different ways,
suggesting that cue-preserving compression requires condition-specific improvements.
Summary showed the strongest tendency to reconstruct the input into an explanatory style.
It substantially decreased conversational hedges while increasing formal hedges,
suggesting that it removed spoken-style uncertainty
and made the input more similar to polished explanatory text.
Because the increase in formal hedges was smaller than the decrease in conversational hedges,
this change should be interpreted mainly as the removal of conversational uncertainty.

RAG appears to reshape the input through retrieval selection rather than by rewriting the entire text.
Unlike Summary, RAG did not show broad changes in J-LIWC categories
or a significant increase in formal hedges.
However, it decreased conversational hedges and increased the overall occurrence rate
of institutional and technical terms.
This suggests that retrieval may remove peripheral context and spoken fluctuations
while making technical and institutional vocabulary more salient.

Screening achieved the classification performance closest to the Baseline among
the compressed inputs, but it also introduced information loss at the extraction
stage. The extraction analysis showed relatively high precision but low recall,
indicating that extracted sentences were often medical or health-related, whereas
many relevant sentences were missed. This pattern suggests that Screening can
reduce redundant speech and preserve some explicit medical claims, but it may
also exclude important surrounding context. In particular, medically relevant
content can be missed when it is embedded in long explanatory sequences,
expressed indirectly through metaphorical language, or distributed across
testimonials and narrative context. Conversely, some non-substantive utterances,
such as greetings, may be extracted as medical or health-related sentences. These
errors indicate that future screening methods should not merely select sentences
containing explicit medical terminology, but should preserve the local context in
which claims, explanations, testimonials, and persuasive cues are presented.

\subsection{Implications and Limitations}

These findings suggest that input compression for medical misinformation detection
should preserve not only claim content
but also style, hedging, affect, and institutional context.
Institutional and technical terms may support valid medical explanations,
but they may also lend an appearance of authority to misleading claims.
Therefore, their foregrounding in compressed inputs does not necessarily indicate medical validity.
Rather, it may function as a surface-level cue that encourages LLMs to judge the input as Real.

Several limitations should be considered
when interpreting these findings, particularly the dataset size, annotation procedure,
and dictionary-based linguistic analysis.
First, the analysis was based on
74 Japanese-language medical YouTube videos,
and the dataset was limited in size and topic coverage.
Because the number of Real videos was small,
future work should verify the generalizability of the findings using larger and more balanced datasets.
Second, the labeling and the evaluation of Screening extraction performance
were conducted by a single author,
leaving room for annotation subjectivity.
Third, the linguistic analysis relied on occurrence rates
based on J-LIWC, a hedge dictionary, and an institutional and technical term dictionary,
and therefore did not fully examine the semantic functions of each expression in context.
Future work should include multiple annotators,
contextual analysis of linguistic expressions, and investigation
of which input segments LLMs use as evidence for classification.

%% file: section/7_conclusion.tex
This study examined how transcript compression affects
LLM-based veracity classification of Japanese medical YouTube videos.
We compared full-transcript input with three compressed input designs:
LLM-generated summaries, RAG-based retrieved contexts,
and extracted candidate medical and health-related sentences.
The full-transcript Baseline achieved the best performance,
whereas all compressed inputs increased false negatives,
indicating that Fake videos became more likely to be misclassified as Real after compression.
Linguistic analyses showed that this tendency could not be explained
by a simple increase in certainty.
Instead, Summary reduced affective, cognitive, temporal, social, and conversational cues,
while Summary and RAG made institutional and technical terms relatively more salient.
These findings suggest that transcript compression can represent Fake videos
as more coherent and authoritative inputs, thereby weakening cues needed
for misinformation detection.

When transcripts fit within the context window and computational cost is acceptable,
full-transcript input may be safer than generic compression methods.
Future work should develop cue-preserving compression methods that reduce input length
while preserving explicit claims as well as the linguistic
and contextual cues surrounding them.

%% file: section/8_acknowledgements.tex
This work was supported by JSPS KAKENHI Grant Numbers JP23K24949 and 25K03105,
and by a joint research project with SKY Inc. (CPI0405).